\documentclass{article}

\usepackage{arxiv}
\usepackage[utf8]{inputenc} 
\usepackage[T1]{fontenc}    
\usepackage{hyperref}       
\usepackage{url}            
\usepackage{booktabs}       
\usepackage{amsfonts}       
\usepackage{nicefrac}       
\usepackage{microtype}      
\usepackage{lipsum}		
\usepackage{graphicx}
\usepackage[numbers, compress]{natbib}
\usepackage{doi}
\usepackage{multirow}
\usepackage{rotating}
\usepackage{diagbox}

\title{Investigation of domain gap problem in several deep-learning-based CT metal artefact reduction methods}

\author{Muge~Du\\
	Department of Engineering Physics\\
	Tsinghua University\\
	\texttt{dmg19@mails.tsinghua.edu.cn} \\
	\And
	Kaichao~Liang \\
	Department of Engineering Physics\\
	Tsinghua University\\
	\And
	Yinong~Liu \\
	Department of Engineering Physics\\
	Tsinghua University\\
	\And
	Yuxiang~Xing* \\
	Department of Engineering Physics\\
	Tsinghua University\\
}

\date{}

\renewcommand{\headeright}{}
\renewcommand{\undertitle}{}
\renewcommand{\shorttitle}{Investigation of domain gap problem in deep CT MAR}

\hypersetup{
pdftitle={Investigation of domain gap problem in several deep-learning-based CT metal artefact reduction methods},
pdfsubject={q-bio.NC, q-bio.QM},
pdfauthor={Muge~Du, Kaichao~Liang, Yinong~Liu, Yuxiang~Xing},
pdfkeywords={Metal Artefact Reduction, CT, Domain Gap, Deep Learning},
}

\begin{document}
\maketitle

\begin{abstract}
Metal artefacts in CT images may disrupt image quality and interfere with diagnosis. Recently many deep-learning-based CT metal artefact reduction (MAR) methods have been proposed. Current deep MAR methods may be troubled with domain gap problem, where methods trained on simulated data cannot perform well on practical data. In this work, we experimentally investigate two image-domain supervised methods, two dual-domain supervised methods and two image-domain unsupervised methods on a dental dataset and a torso dataset, to explore whether domain gap problem exists or is overcome. We find that I-DL-MAR and DudoNet are effective for practical data of the torso dataset, indicating the domain gap problem is solved. However, none of the investigated methods perform satisfactorily on practical data of the dental dataset. Based on the experimental results, we further analyze the causes of domain gap problem for each method and dataset, which may be beneficial for improving existing methods or designing new ones. The findings suggest that the domain gap problem in deep MAR methods remains to be addressed.
\end{abstract}


\section{Introduction}
Metals contained in objects scanned by CT will cause metal artefacts in reconstructed images that appear in various form of bright or dark streaking, banding and shading artefacts \citep{barrett_review_2004,gjesteby_review_2016} and may interfere with the diagnosis. Multiple metal artefact reduction (MAR) methods have been proposed to eliminate the influence of metal artefacts. Among them, recently proposed deep-learning-based MAR (referred as deep MAR) methods show more promising performance than traditional MAR methods.

Existing deep MAR methods are mostly based on supervised learning, which requires the pairs of artefact-affected data and metal-free label for training, thus a distance function between network outputs and corresponding labels can be minimized. These supervised MAR methods can be divided into three categories: image-domain \citep{gjesteby_id_2017,huang_idresnet_2018,zhu_idunet_2019,wang_idgan_2019,fan_idquad_2019,liang_idlmar_2019}, sinogram-domain \citep{park_sd_2018,ghani_sd_2019,liao_sdgan_2019}, and dual-domain \citep{zhang_cnnmar_2018,lin_dudonet_2019,lyu_dudonet++_2020,peng_dd_2020,wang_dd_idol-net_2021,yu_ddres_2021}. Image domain supervised MAR methods work purely on reconstructed images and adopt techniques including residual learning \citep{huang_idresnet_2018,fan_idquad_2019}, advanced network designs \citep{zhu_idunet_2019, wang_idgan_2019} like U-Net \citep{ronneberger_unet_2015} and conditional GAN \citep{isola_pix2pix_2017} to improve the performance. Sinogram-domain deep MAR methods learn to reduce artefacts by processing the projections, where the corrupted signals inside metal traces are rectified. Dual-domain MAR methods reduce metal artefacts by recovering the information in both sinogram and image domain. Representative dual-domain methods include the earlier CNNMAR \citep{zhang_cnnmar_2018} which combines conventional NMAR \citep{meyer_nmar_2010} in sinogram domain and deep MAR in image domain, and later methods like DudoNet \citep{lin_dudonet_2019} and DudoNet++ \citep{lyu_dudonet++_2020} which jointly train the sinogram-domain and image-domain networks and connect them with differentiable reconstruction layer and can recover certain structures whose information is submerged in single domain. Dual-domain methods usually perform better compared to single-domain methods in simulation studies.

Since it is hard to obtain paired with/without artefact data in practice, most supervised MAR methods are trained on simulated dataset composed of pairs of simulated artefact-affected data and artefact-free labels. However, as the simulation of metal artefacts is often imperfect and cannot precisely capture the actual physics causing artefacts, difference between simulated artefacts and practical artefacts is unavoidable. Such difference will cause the supervised MAR method trained on simulated dataset (source domain) to perform poorly on practical dataset (target domain), which could be referred as the domain gap problem.

To alleviate domain gap problem in MAR, researchers have proposed several unsupervised MAR methods which directly learn the MAR task on practical artefact-affected data and unpaired metal-free data, usually by a GAN-based image-to-image translation framework called CycleGAN \citep{zhu_cyclegan_2017}. \citep{du_cyclegan_2018} first applies CycleGAN to fan beam dental CT MAR and demonstrates its feasibility to solve domain gap problem. \citep{nakao_3dcyclegan_2020} extends CycleGAN from 2D to 3D for better reduction of metal artefacts and introduces extra regularizations to avoid distortion of anatomical structures. \citep{lee_betacyclegan_2021} proposes $\beta$-CycleGAN-based MAR with attention mechanism. ADN \citep{liao_adn_2020} proposes to disentangle the metal artefacts and contents in encoded feature space based on a GAN-based framework similar to CycleGAN. ADN is later improved by adding low-dimensional manifold
 constraints \citep{niu_adnldm_2020} or using multimodal information from CT and MRI \citep{ranzini_adnmultimodal_2020}. Besides above image-domain unsupervised methods, recently \citep{lyu_cycledudonet_2021} proposes an dual-domain unsupervised method based on CycleGAN. These unsupervised MAR methods have shown effectiveness of reducing practical metal artefacts in experimental studies on certain datasets. However, these methods may face more difficulty in training compared to supervised methods due to the well-known instability of GAN \citep{goodfellow_gan_2014} used in these methods. It is unclear whether the performance of unsupervised MAR methods is always good or depends on certain datasets.

In this work, we investigate the domain gap problem in several deep MAR methods, which is crucial for clinical application. The investigated methods cover image-domain supervised MAR (a supervised baseline and I-DL-MAR \citep{liang_idlmar_2019}), dual-domain supervised MAR (DudoNet \citep{lin_dudonet_2019} and DudoNet++ \citep{lyu_dudonet++_2020}) and image-domain unsupervised MAR (CycleGAN-based MAR \citep{du_cyclegan_2018, nakao_3dcyclegan_2020} and ADN \citep{liao_adn_2020}). The comparison is done on a dental dataset and a torso dataset which both contain simulated and practical data. Compared to previous clinical study for evaluating unsupervised MAR methods \citep{du_cyclegan_2018, liao_adn_2020}, the two datasets are of higher resolution, and especially the dental dataset contains more severe artefacts. In addition, we analyze the causes of domain gap problem in each method based on the experimental results and give possible recommendations. The findings of this work could help understand how to avoid domain gap problem in existed MAR methods and design new methods.

\section{Investigated MAR Methods}
\label{sec:methods}
In this work, six existing deep MAR methods are investigated, including four supervised methods with two in image domain and the other two in dual domain, and two unsupervised methods in image-domain. In the following, the notations in each method have been rewritten for consistency.

The notations in this work are: $\mathrm{S}$ denotes the source domain, defined as the data affected by simulated artefacts, while $\mathrm{T}$ denotes the target domain, defined as the data affected by practical artefacts. $\upmu$ denotes the CT image, and $\mathbf{p}$ denotes the sinogram. Superscription $\mathrm{fr}$ denotes the data is artefact-free, superscription $\mathrm{ma}$ denotes the data is affected by metal artefacts, and superscription $\mathrm{LI}$ denotes the data is linear-interpolated.

\subsection{Image-domain supervised MAR methods}
Two image-domain supervised MAR methods are investigated: a common supervised MAR method referred as the supervised baseline, and I-DL-MAR.
\paragraph{The supervised baseline}
In the supervised baseline, a deep MAR network $\mathcal{R}$ takes the original artefact-affected CT image $\upmu^{\mathrm{ma}}$ as input and estimates an artefact-reduced image ${\hat{\upmu}}^\mathrm{fr}$:
\begin{equation}
    {\hat{\upmu}}^\mathrm{fr}=\mathcal{R}(\upmu^{\mathrm{ma}})
\end{equation}
During training, only source domain data is used. The loss function is defined as the L1 norm between the non-metal region of the artefact-reduced estimation ${\hat{\upmu}}_\mathrm{S}^\mathrm{fr}$ and corresponding artefact-free labels ${\upmu}_\mathrm{S}^\mathrm{fr}$, formulated as:
\begin{equation}
    \mathcal{L}_\text{supervised baseline}=\mathbb{E}_{{\upmu}_\mathrm{S}^\mathrm{ma}}\lVert ({\hat{\upmu}}_\mathrm{S}^\mathrm{fr}-{\upmu}_\mathrm{S}^\mathrm{fr}) \odot (1-\mathbf{M}({{\upmu}_\mathrm{S}^\mathrm{ma}}))\rVert=\mathbb{E}_{{\upmu}_\mathrm{S}^\mathrm{ma}}\lVert (\mathcal{R}(\upmu_\mathrm{S}^{\mathrm{ma}})-{\upmu}_\mathrm{S}^\mathrm{fr}) \odot (1-\mathbf{M}({{\upmu}_\mathrm{S}^\mathrm{ma}}))\rVert \label{eqt:supervised}
\end{equation}
where $\mathbb{E}_x$ denotes the expectation over $x$, $\mathbf{M}(\upmu)$ is the binary mask of the metal region in the input $\upmu$, and $\odot$ denotes the element-wise multiplication.

\paragraph{I-DL-MAR}
I-DL-MAR \citep{liang_idlmar_2019} is a supervised method proposed for reducing metal artefacts in practical CT images. It modifies the supervised baseline by taking the linear-interpolated (LI) image $\upmu^{\mathrm{LI}}$ as input, instead of $\upmu^{\mathrm{ma}}$:
\begin{equation}
    {\hat{\upmu}}^\mathrm{fr}=\mathcal{R}(\upmu^{\mathrm{LI}})
\end{equation}
The loss function for I-DL-MAR is similar to \eqref{eqt:supervised}, and only differs in replacing input from $\upmu^{\mathrm{ma}}$ to $\upmu^{\mathrm{LI}}$:
\begin{equation}
    \mathcal{L}_\text{I-DL-MAR}=\mathbb{E}_{{\upmu}_\mathrm{S}^\mathrm{LI}}\lVert ({\hat{\upmu}}_\mathrm{S}^\mathrm{fr}-{\upmu}_\mathrm{S}^\mathrm{fr}) \odot (1-\mathbf{M}({{\upmu}_\mathrm{S}^\mathrm{ma}}))\rVert = \mathbb{E}_{{\upmu}_\mathrm{S}^\mathrm{LI}}\lVert (\mathcal{R}(\upmu_\mathrm{S}^{\mathrm{LI}})-{\upmu}_\mathrm{S}^\mathrm{fr}) \odot (1-\mathbf{M}({{\upmu}_\mathrm{S}^\mathrm{ma}}))\rVert_1 \label{eqt:I-DL-MAR}
\end{equation}
Ideally, to acquire the input ${\upmu}^\mathrm{LI}$, the domain-variant metal-affected signals in projections are replaced with linear interpolation, thus ${\upmu}^\mathrm{LI}$ is nearly domain-invariant, and I-DL-MAR can avoid domain gap problem.

\subsection{Dual-domain supervised MAR methods}
Two supervised dual-domain MAR methods are investigated: DudoNet \citep{lin_dudonet_2019} and DudoNet++ \citep{lyu_dudonet++_2020}. In these two methods, metal artefacts are reduced by a sinogram enhancement network SE-Net $\mathcal{R}_\mathrm{SE}$ and an image enhancement network IE-Net $\mathcal{R}_\mathrm{IE}$. 

Given the corrupted input sinogram $\mathbf{p}$ and its metal trace $\mathbf{M}_t\left(\mathbf{p}^{\mathrm{ma}}\right)$, $\mathcal{R}_\mathrm{SE}$ corrects the signal only inside the metal trace, resulting in the predicted sinogram ${\hat{\mathbf{p}}}^{\mathrm{fr},\mathrm{SE}}$.
\begin{equation}
    {\hat{\mathbf{p}}}^{\mathrm{fr},\mathrm{SE}}=\mathcal{R}_\mathrm{SE}(\mathbf{p},\mathbf{M}_t\left(\mathbf{p}\right)) \odot \mathbf{M}_t\left(\mathbf{p}^{\mathrm{ma}}\right) + \mathbf{p} \odot (1-\mathbf{M}_t\left(\mathbf{p}^{\mathrm{ma}}\right))
\end{equation}
Then a differentiable Radon Inversion Layer (RIL) $\mathcal{P}^{-1}$ reconstructs the former corrected sinogram into image ${\hat{\upmu}}^{\mathrm{fr},\mathrm{SE}}$.
\begin{equation}
    {\hat{\upmu}}^{\mathrm{fr},\mathrm{SE}}=\mathcal{P}^{-1}({\hat{\mathbf{p}}}^{\mathrm{fr},\mathrm{SE}})
\end{equation}
Finally, $\mathcal{R}_\mathrm{IE}$ takes ${\hat{\upmu}}^{\mathrm{fr},\mathrm{SE}}$ and the corrupted image $\upmu$ as input and predicts the corrected artefact-free image ${\hat{\upmu}}^{\mathrm{fr},\mathrm{IE}}$. Specifically, in DudoNet++, the image-domain metal mask is an additional input to $\mathcal{R}_\mathrm{IE}$.
\begin{equation}
    {\hat{\upmu}}^{\mathrm{fr},\mathrm{IE}}=\begin{cases}
    \mathcal{R}_\mathrm{IE}({\hat{\upmu}}^{\mathrm{fr},\mathrm{SE}},\upmu) & \text{for DudoNet}\\
    \mathcal{R}_\mathrm{IE}({\hat{\upmu}}^{\mathrm{fr},\mathrm{SE}},\upmu, \mathbf{M}(\upmu^\mathrm{ma})) & \text{for DudoNet++}\\
    \end{cases}\label{eqt:IENet}
\end{equation}
DudoNet and DudoNet++ are also trained only on source domain. There are three supervised loss functions, related to the fidelity of SE-Net, reconstruction consistency and the fidelity of IE-Net, defined by comparing the result with corresponding label. Specifically, the loss for IE-Net in DudoNet is computed on the whole image, while in DudoNet++ is computed on the  non-metal region.
\begin{equation}
    \mathcal{L}_\text{SE}=\mathbb{E}_{\mathbf{p}_\mathrm{S}}\lVert ({\hat{\upmu}}_\mathrm{S}^{\mathrm{fr},\mathrm{SE}} - \mathbf{p}_\mathrm{S}^\mathrm{fr}\rVert_1 \label{eqt:SELoss}
\end{equation}
\begin{equation}
    \mathcal{L}_\text{RC}=\mathbb{E}_{\mathbf{p}_\mathrm{S}}\lVert ({\hat{\upmu}}_\mathrm{S}^{\mathrm{fr},\mathrm{SE}} - \upmu_\mathrm{S}^\mathrm{fr}) \odot (1-\mathrm{M}(\upmu_\mathrm{S}^\mathrm{ma} ))\rVert_1 \label{eqt:RCLoss}
\end{equation}
\begin{equation}
     \mathcal{L}_\text{IE}=\begin{cases}
     \mathbb{E}_{\mathbf{p}_\mathrm{S}}\lVert{\hat{\upmu}}_\mathrm{S}^{\mathrm{fr},\mathrm{IE}}-\upmu_\mathrm{S}^\mathrm{fr}\rVert_1 & \text{for DudoNet}\\
     \mathbb{E}_{\mathbf{p}_\mathrm{S}}\lVert({\hat{\upmu}}_\mathrm{S}^{\mathrm{fr},\mathrm{IE}}-\upmu_\mathrm{S}^\mathrm{fr}) \odot (1-\mathrm{M}(\upmu_\mathrm{S}^\mathrm{ma}))\rVert_1 & \text{for DudoNet++}\\
     \end{cases} \label{eqt:IELoss}
\end{equation}
DudoNet++ is different from DudoNet in several aspects.

1. As for the corrupted sinogram $\mathbf{p}$ and image $\upmu$ input to the method, DudoNet takes the linear-interpolated data $\mathbf{p}^\mathrm{LI}$ and $\upmu^\mathrm{LI}$ as input, while DudoNet++ takes the original metal-artefact-affected data $\mathbf{p}^\mathrm{ma}$ and $\upmu^\mathrm{ma}$ as input.

2. As for IE-Net $\mathcal{R}_\mathrm{IE}$, compared to DudoNet, DudoNet++ additionally utilizes the image-domain metal mask as both the input to the network and as the component of the IE loss function, as shown in \eqref{eqt:IENet} and \eqref{eqt:IELoss}.

3. Improvements of sinogram padding strategy and network structures are also adopted in DudoNet++. The details of these improvements are omitted here and can be found in \citep{lyu_dudonet++_2020}.

Dual domain methods can utilize information from both sinogram and image domains, which benefits the recovery of corrupted signals compared to image-domain methods. Among these two methods, DudoNet is fed with the LI data which is often less domain-variant but may discard useful information during interpolation, while DudoNet++ is fed with the original artefact-affected data which contains complete information but may be more prone to domain gap problem.

\subsection{Image-domain unsupervised MAR methods}
Two unsupervised MAR methods are investigated: an improved version of CycleGAN-based MAR \citep{du_cyclegan_2018} and ADN \citep{liao_adn_2020}. They are representative unsupervised MAR methods and can be trained with unpaired practical data.
\paragraph{CycleGAN-INT}
CycleGAN \citep{zhu_cyclegan_2017} learns a cyclic unpaired mapping between two domains and has been applied to 2D and 3D MAR problem, respectively. CycleGAN-based MAR has two GANs, each is composed of a generator and a discriminator: $(G^\mathrm{fr},\ D^\mathrm{fr})$ and $(G^\mathrm{ma},\ D^\mathrm{ma})$. $G^\mathrm{fr}$ learns to reduce metal artefacts from $\upmu^\mathrm{ma}$, while $G^\mathrm{ma}$ learns to add metal artefacts to $\upmu^{\mathrm{fr}\prime}$. $\upmu^\mathrm{ma}$ and $\upmu^{\mathrm{fr}\prime}$ are unpaired, thus allowing using practical artefact-affected images for training.
\begin{equation}
    \hat{\upmu}^\mathrm{fr} = G^\mathrm{fr}(\upmu^\mathrm{ma})
\end{equation}
\begin{equation}
    \hat{\upmu}^{\mathrm{ma}\prime} = G^{\mathrm{ma}}(\upmu^{\mathrm{fr}\prime})
\end{equation}
CycleGAN-based MAR trains two GANs with common adversarial training and additional constraints. As for the specific adversarial loss $\mathcal{L}_\text{adv}$ for training GAN, CycleGAN-MAR uses LS-GAN framework \citep{mao_lsgan_2017}, whose detailed formulas are omitted here. As for losses of additional constraints, the basic one is cycle consistency loss, which requires the processing of each generator can be mostly reversed by the other generator. The cycle consistency loss $\mathcal{L}_\text{cyc}$ is defined below:
\begin{equation}
    \mathcal{L}_\text{cyc, ma}=\mathbb{E}_{\upmu^\mathrm{ma}}\lVert G^\mathrm{ma}(\hat{\upmu}^\mathrm{fr}) - \upmu^\mathrm{ma} \rVert_1=\mathbb{E}_{\upmu^\mathrm{ma}}\lVert G^\mathrm{ma}(G^\mathrm{fr}(\upmu^\mathrm{ma})) - \upmu^\mathrm{ma} \rVert_1
\end{equation}
\begin{equation}
    \mathcal{L}_\text{cyc, fr}=\mathbb{E}_{\upmu^{\mathrm{fr}\prime}}\lVert G^\mathrm{fr}(\hat{\upmu}^\mathrm{ma}) - \upmu^{\mathrm{fr}\prime} \rVert_1=\mathbb{E}_{\upmu^{\mathrm{fr}\prime}}\lVert G^\mathrm{fr}(G^\mathrm{ma}(\upmu^{\mathrm{fr}\prime})) - \upmu^{\mathrm{fr}\prime} \rVert_1
\end{equation}
\begin{equation}
    \mathcal{L}_\text{cyc} = \mathcal{L}_\text{cyc, ma} + \mathcal{L}_\text{cyc, fr}
\end{equation}
Another frequently used loss is identity loss $\mathcal{L}_\text{idt}$ defined as
\begin{equation}
    \mathcal{L}_\text{idt}=\mathbb{E}_{\upmu^\mathrm{ma}}\lVert G^\mathrm{ma}(\upmu^\mathrm{ma}) - \upmu^\mathrm{ma} \rVert_1 + \mathbb{E}_{\upmu^{\mathrm{fr}\prime}}\lVert G^\mathrm{fr}(\upmu^{\mathrm{fr}\prime}) - \upmu^{\mathrm{fr}\prime} \rVert_1
\end{equation}
Above are the fundamental losses of CycleGAN-based MAR. Recently, the possibility of changing anatomic structures in the result of CycleGAN-based MAR has been reported \citep{nakao_3dcyclegan_2020} and an intensity loss is introduced in \citep{nakao_3dcyclegan_2020} to discourage the distortion of structures. We add it into the implementation of CycleGAN-based MAR and refer the improved version as CycleGAN-INT. The intensity loss $\mathcal{L}_\text{int}$ is formulated as:
\begin{equation}
    \mathcal{L}_\text{int}=\mathbb{E}_{\upmu^\mathrm{ma}}\lVert G^\mathrm{fr}(\upmu^\mathrm{ma}) - \upmu^\mathrm{ma} \rVert_1 + \mathbb{E}_{\upmu^{\mathrm{fr}\prime}}\lVert G^\mathrm{ma}(\upmu^{\mathrm{fr}\prime}) - \upmu^{\mathrm{fr}\prime} \rVert_1
\end{equation}
For CycleGAN-INT, $\mathcal{L}_\text{adv}$, $\mathcal{L}_\text{cyc}$, $\mathcal{L}_\text{idt}$ and $\mathcal{L}_\text{int}$ are optimized together during training.
\paragraph{ADN}
ADN \citep{liao_adn_2020} is another unsupervised image-domain MAR method that encourages the disentanglement of metal artefacts and contents. Like CycleGAN, ADN also trains two GANs to perform artefact reduction and artefact generation, and the training is also under additional constraints including cycle consistency. The data-processing and the loss functions in ADN are very complex, thus we only summarize its features here and omit the detailed notations and formulas.

Different from CycleGAN where each generator is a whole encoder-decoder network, ADN uses separated encoders and decoders that allow flexibility in recompositing features and generating images. There are three encoders in ADN, two for extracting artefacts and contents from the artefact-affected image, and one for extracting contents from the artefact-free image. There are two decoders for reconstructing artefact-affected and artefact-free images from the encoded features. 

The constraints to train ADN are similar to CycleGAN but are also designed to utilize the flexibility of separate encoders and decoders and to fulfill the purpose of disentanglement. It also introduces an artefact-consistency loss which requires the generated and the reduced artefacts in decoded images to be close if the artefacts are from the same encoded artefact feature.
\section{Experiments}
\label{sec:experiments}
\subsection{Configurations}
\subsubsection{Datasets}
In this work, we evaluate the performance of MAR on a dental CT image dataset and a torso CT image dataset. All raw datasets contain clinical metal-inserted and metal-free images of different patients. After processing, each dataset is composed of a source domain sub-dataset containing simulation artefact-affected data, and a target domain sub-dataset containing practical artefact-affected data.
\paragraph{Dental dataset}
The dental dataset is built based on raw data acquired from a commercial cone-beam dental CT. It has a half-detector projection with 656 detectors and collects data of 600 views over $2\pi$. The reconstructed image is $640\times640$ with pixel size 0.25mm. As we conduct our study on 2D image, the projection simulation and the artefact simulation use a fan-beam CT configuration, same to the middle-slice of the original cone-beam CT.

For source domain sub-dataset, we use the simulation method in \citep{liang_idlmar_2019} to insert metals ($50\%$ Ti and $50\%$ Fe) into real metal-free CT slices and simulate metal artefacts. In short, for a metal-free dental CT slice, we randomly select one or more teeth and insert practical metal implants cropped from real patients into the teeth region, and manually refine the inserted metal mask to be more realistic if needed. The teeth region can be completely or partially replaced according to practical situation. The artefact-affected projection is then simulated from the metal-inserted image using a polychromatic ray-tracing model with additional noise. FBP is used for reconstruction. In total, the resulted source domain sub-dataset has 2704 pairs of simulated artefact-affected data and artefact-free data.

To form the target domain sub-dataset, we select 1601 clinical artefact-affected images with maximum pixel value above 5000HU.

Both source domain sub-dataset and target domain sub-dataset are split into training set, validation set, and test set by the ratio of 7:2:1.
\paragraph{Torso dataset}
The torso dataset is built based on raw data from an open-source dataset DeepLesion \citep{yan_deeplesion_2018}. As DeepLesion provides no CT projection for each image, we simulate the projection with 640 views and 642 detectors using a ray-tracing model. The resized image size and the reconstruction size are set to $416\times416$, same to the configuration used for DeepLesion in the papers of the investigated methods DudoNet\citep{lin_dudonet_2019} and DudoNet++\citep{lyu_dudonet++_2020} for fair comparison, and is larger than the configuration in ADN\citep{liao_adn_2020}.

For source domain sub-dataset, we simulate artefact-affected images from the metal-free background images in DeepLesion. We adopt the same metal artefact simulation workflow proposed in \citep{liao_adn_2020}. 4118 metal-free images are randomly selected from DeepLesion as the label, where 3918 are for training, and 200 are for testing. Each training label is assigned with 90 fixed metal masks, and each testing label is assigned with other 10 masks. Metal material is set to Ti for both training and testing. The metal artefact simulation method proposed by \citep{zhang_cnnmar_2018} is used to simulate artefact-affected images.

To form the target domain sub-dataset, we select 1583 artefact-affected clinical images from 173 scans of 74 patients from DeepLesion. All the images are with more than 100 pixels above 3000 HU. Among them, 60 patients are used for training, and the remaining 14 patients are used for testing.
\subsubsection{Implementation details}
\paragraph{Network Structure}
The MAR networks used for supervised baseline, I-DL-MAR and CycleGAN-Int are the 4-layer U-Net \citep{ronneberger_unet_2015}, which has 5 encoding blocks (considering the bottleneck is the fifth) and 4 decoding blocks, with each block containing 2 CNN layers. The DudoNet and DudoNet++ adopt their original U-Net-based structures, and ADN uses its original encoder-decoder structure. The discriminator in CycleGAN-Int and ADN are the patch-discriminator. The minimal unit number of network channels in all the methods is 32, except in ADN which is 64.
\paragraph{Code Implementation}
All the methods are implemented in PyTorch. ADN are implemented based on its official open-source code\footnote{https://github.com/liaohaofu/adn}. CycleGAN-INT is implemented with open-source code of CycleGAN\footnote{https://github.com/junyanz/pytorch-CycleGAN-and-pix2pix}. We implement the remaining methods by ourselves as no official code is available.
\paragraph{Training Hyperparameters}
All the methods are trained with Adam optimizer with $(\beta_1, \beta_2)$ set to (0.5, 0.999). For ADN, DudoNet and DudoNet++, we follow the same learning rate initialization and adjustment strategy as in their original papers. For other methods, the learning rate is fixed at 0.0002 during the first $70\%$ epochs and is linearly decayed to zero during the remaining $30\%$ epochs. The batch size is two for ADN\footnote{In our experience, ADN trained with large batch size (4 and 8) performs worse than ADN trained with batch size 2.} and eight for other methods.

For CycleGAN-INT, the weights of cycle loss, identity loss, intensity loss and adversarial loss are 10, 5, 25, and 1 respectively, following the recommended setting in \citep{nakao_3dcyclegan_2020}. For ADN, the weights of reconstruction loss, self-reduction loss, and artefact consistency loss are 20, and the weight of adversarial loss is 1, following the hyper-parameter setting for dataset SYN and CL1 in \citep{liao_adn_2020}. For DudoNet and DudoNet++, we follow their default hyper-parameter setting where all the loss functions are equally weighted.
\paragraph{Other details}
Supervised methods are trained using simulated data that have paired input and labels. While for unsupervised methods, training data is determined according to the type of artefacts in evaluation. I.e. for clinical study, CycleGAN-INT and ADN are trained using practical artefact-affected images from the target domain. While for simulation study, they are trained using simulated artefact-affected images from the source domain. To ensure that when training unsupervised methods on simulated data, the artefact-free data has no overlapped information with the artefact-affected data, we adopt the dataset-splitting strategy in ADN, where the whole simulated training dataset is 1:1 divided, half used for artefact-affected data, the other half used for artefact-free data.

\subsection{Simulation Study of MAR}
\label{sec:simulation}
We first report the results of simulation study on source domain for both dental and torso datasets. The performance on source domain measures the basic ability of MAR, especially for supervised methods.

\begin{table}
	\caption{Quantitative evaluation results on source domain for: (a) dental dataset and (b) Torso dataset. ID means image-domain, DD means dual-domain. For each metric, the best result among all methods is shown in bold.}
	\centering
	\resizebox{\linewidth}{!}{
	\renewcommand{\arraystretch}{1.2}
	\begin{tabular}{cc|cccccccc}
        \toprule
        \multicolumn{2}{c}{ \multirow{2}*{} } &\multicolumn{2}{c}{} &\multicolumn{2}{c}{ID Supervised} &\multicolumn{2}{c}{DD Supervised}  &\multicolumn{2}{c}{ID Unsupervised}\\
        \cline{5-10}
        \multicolumn{2}{c}{} &Input(MA) & LI & {Supervised baseline}& I-DL-MAR & DudoNet & DudoNet++ & CycleGAN-INT & ADN\\
        \hline
        \multirow{3}*{\rotatebox{90}{(a) dental}}
        &RRMSE & 0.1904 & 0.1046 & 0.0560 & 0.0556 & 0.0549 & \textbf{0.0431} & 0.0958 & 0.1119\\
        &SSIM & 0.8737 & 0.9148 & 0.9419 & 0.9548 & 0.9684 & \textbf{0.9737} & 0.8964 & 0.8514\\
        &PSNR & 28.7487 & 33.9742 & 39.0253 & 39.1531 & 39.4387 & \textbf{41.6102} & 34.5396 & 33.1685\\
        \hline
        \multirow{3}*{\rotatebox{90}{(b) torso}}
        &RRMSE & 0.1926 & 0.0679 & 0.0266 & 0.0277 & 0.0269 & \textbf{0.0218} & 0.0814 & 0.0646\\
        &SSIM & 0.7065 & 0.9564 & 0.9901 & 0.9920 & 0.9927 & \textbf{0.9933} & 0.9639 & 0.9679\\
        &PSNR & 26.9202 & 36.1979 & 44.4017 & 44.3377 & 44.7335 & \textbf{46.0184} & 34.5018 & 36.8930\\
        \bottomrule
    \end{tabular}
    }
	\label{tab:sourceresult}
\end{table}

The quantitative evaluation are conducted with RRMSE, SSIM and PSNR calculated in the non-metal regions. The images are translated to the attenuation coefficient before evaluation, and the dynamic range to compute SSIM and PSNR is 0.1 for dental dataset, and 0.48 for torso dataset. The results are listed in Table \ref{tab:sourceresult}. From Table \ref{tab:sourceresult}, we can see that dual-domain supervised methods are better than image-domain supervised methods, while the latter are much better than image-domain unsupervised methods. More specifically, DudoNet++ achieves better metrics than DudoNet in both datasets. I-DL-MAR is slightly better than the supervised baseline in dental dataset, but its RRMSE and PSNR is worse in torso dataset. CycleGAN-INT is better than ADN in dental dataset, but worse in torso dataset.

The visual results on source domain for dental and torso datasets are shown in Figure \ref{fig:source_dental} and Figure \ref{fig:source_torso}. We observe that:

\begin{figure}
    \centering
    \includegraphics{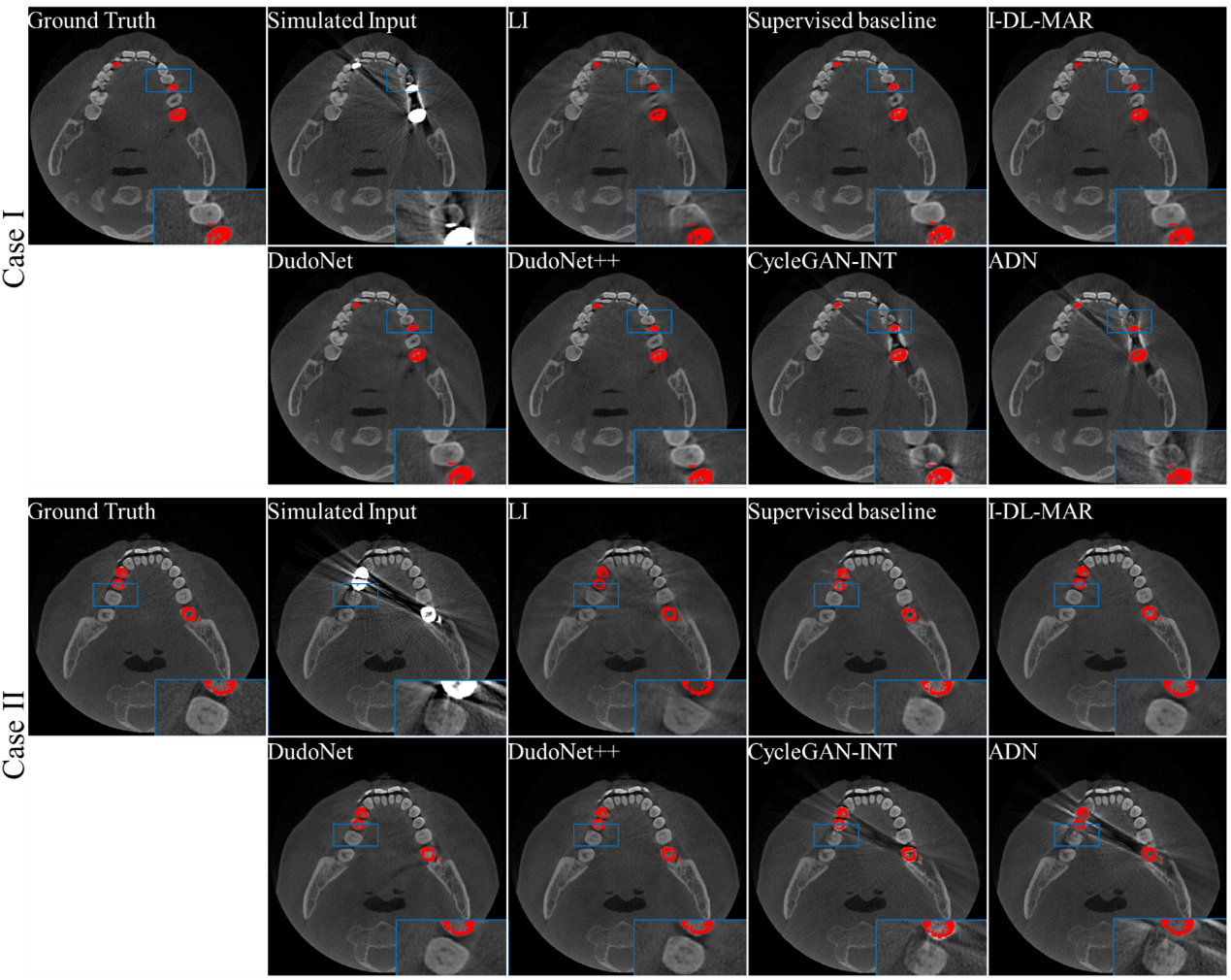}
    \caption{Visual results of two example simulated cases in dental dataset. In each image, area in the blue square is zoomed in for better visualization. For figures in this work, the segmented metals are marked in red.}
    \label{fig:source_dental}
\end{figure}

\begin{figure}
    \centering
    \includegraphics{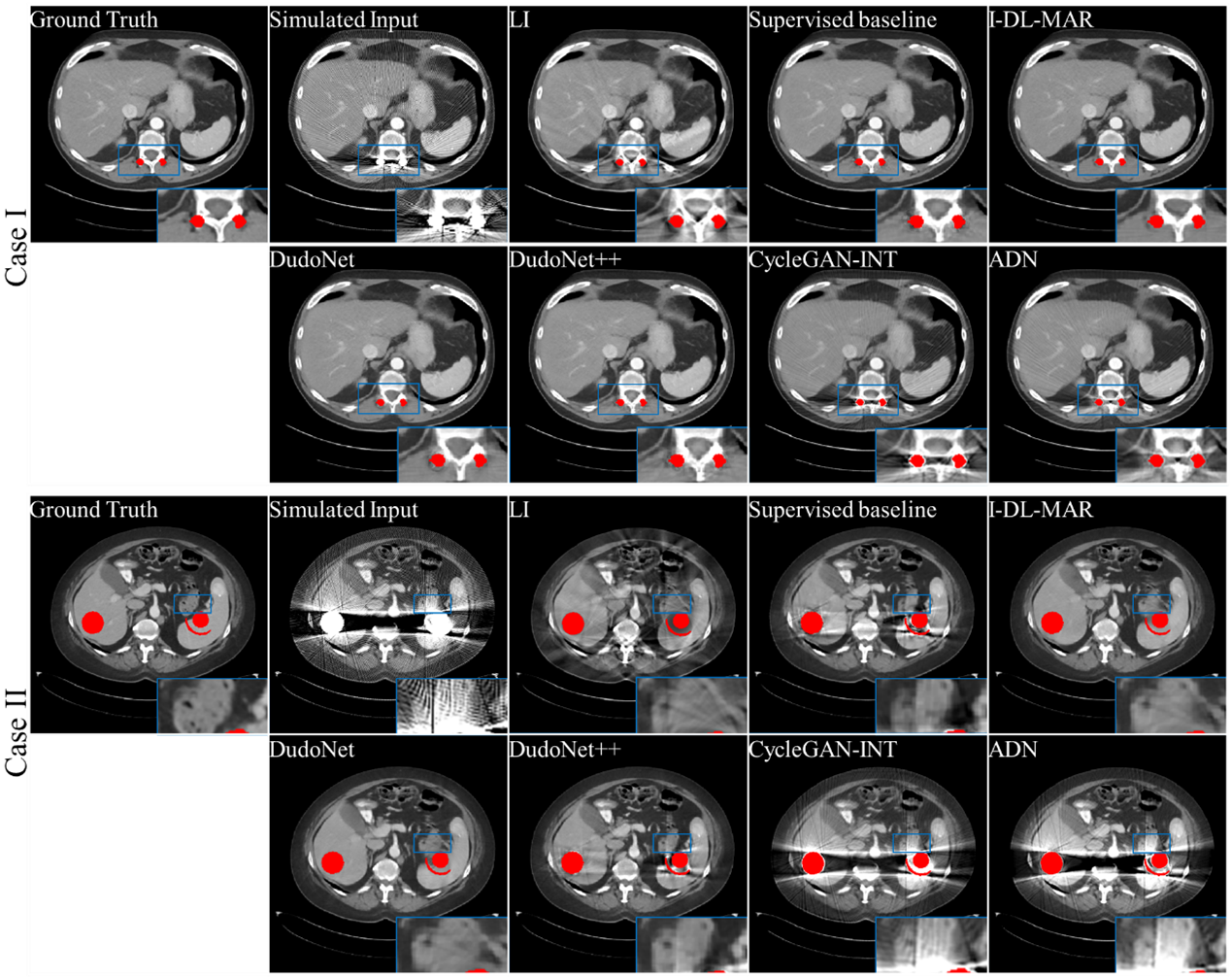}
    \caption{Visual results of two example simulated cases in torso dataset. In each image, area in the blue square is zoomed in for better visualization. For figures in this work, the segmented metals are marked in red.}
    \label{fig:source_torso}
\end{figure}

\paragraph{Dental dataset}

In the two example cases shown in Figure  \ref{fig:source_dental}, simulated input contains strong dark banding artefacts, shading artefacts and streaking artefacts caused by multiple metal implants.

Simply applying LI can reduce most artefacts, especially dark banding artefacts. However, LI introduces new interpolation-based artefacts, and may blur or distort diagnostic structures near metal implants, as shown in the zooming area in two cases.

The supervised baseline can reduce almost all metal artefacts, indicating its effectiveness on source domain. It may be less effective to recover structures covered by heavy dark banding artefacts, as shown in Case I, where the tooth between the two lower metals is imperfectly recovered. On the other hand, as it uses the original data as input, it can maintain the structure near metal implants.

Compared to the supervised baseline, I-DL-MAR gives visually more pleasant results, with cleaner images and less remaining artefacts. It is particularly useful to reduce dark banding artefacts, as shown in Case I, where the teeth between two metals is closer to the ground truth compared to the supervised baseline. Due to the use of LI as input, it inherits and refines the prediction made by LI, which often includes blurred structures. From the zooming area in two cases, we can see that diagnostic structures in I-DL-MAR are mostly maintained. However, in the zooming area in Case I, the tooth shape is different from the ground-truth, indicating I-DL-MAR may have problem of blurred or distorted structures caused by LI.

DudoNet has the result visually similar to I-DL-MAR, since they both use LI data as input. It can reduce most artefacts, but may lead to same structural distortion as I-DL-MAR, shown in the zooming area in Case I. On the other hand, the result of DudoNet++ is visually similar to the supervised baseline, as the two methods both use the original artefact-affected data as input. DudoNet++ is able to reduce artefacts and maintain structures.

The results of image-domain unsupervised methods are visually the worst. The subtle or moderate streaking artefacts are reduced to a certain level and the banding artefacts are alleviated. However, considerable artefacts are remained. CycleGAN-INT performs slightly better than ADN, but far worse than image-domain supervised methods.

\paragraph{Torso dataset}
Compared to dental cases, the streaking artefacts in the two example torso cases shown in Figure \ref{fig:source_torso} are denser and propagated to a wider range of the image, and the dark banding artefacts are thicker due to different parameters for simulation.

LI is still able to reduce most artefacts including dark banding artefacts. However, newly introduced interpolation-based artefacts become stronger in torso dataset and influence a larger area. The structure distortion problem also exists, as shown in the zooming area in Case II.

The supervised baseline is able to completely remove moderate artefacts as shown in Case I. For more severe artefacts in Case II, the supervised baseline can still reduce most artefacts and recover the diagnostic structures well. However, a small portion of artefacts are remained. It is also one of the two best methods in terms of recovering the structures inside the zooming area of Case II (the other method is DudoNet++).

I-DL-MAR again achieves better artefact reduction than the supervised baseline, with almost no artefacts remained in two cases. However, it may blur the structures near large metals, shown in the zooming area of Case II.

DudoNet is similar to I-DL-MAR in terms of both artefact reduction effectiveness and structural loss. The visual difference between the results of two methods is subtle. DudoNet++ behaves similar to the supervised baseline. For stronger artefacts in Case II, DudoNet++ shows slightly better performance than the supervised baseline.

The image-domain unsupervised methods still behave the worst. They are only effective to moderate streaking artefacts.

In summary, on source domain which contains simulated artefacts, both dual-domain supervised methods and image-domain supervised methods are effective. Dual-domain methods are quantitatively the best. Visually, dual-domain methods behave similar to the image-domain methods that share same kind of input data. For image-domain unsupervised methods trained and tested on source domain, their performance are quantitatively and visually the worst.

\subsection{Clinical Study of MAR}
\label{sec:clinical}
We conduct clinical study to evaluate these methods on target domain. For supervised methods whose effectiveness on source domain has been demonstrated, the domain gap exists if the performance drops on target domain. While for unsupervised methods, in clinical study they are trained and tested both on target domain, therefore the result mainly reflects their ability as a solution to domain gap problem by directly learning on target domain.

\begin{figure}
    \centering
    \includegraphics{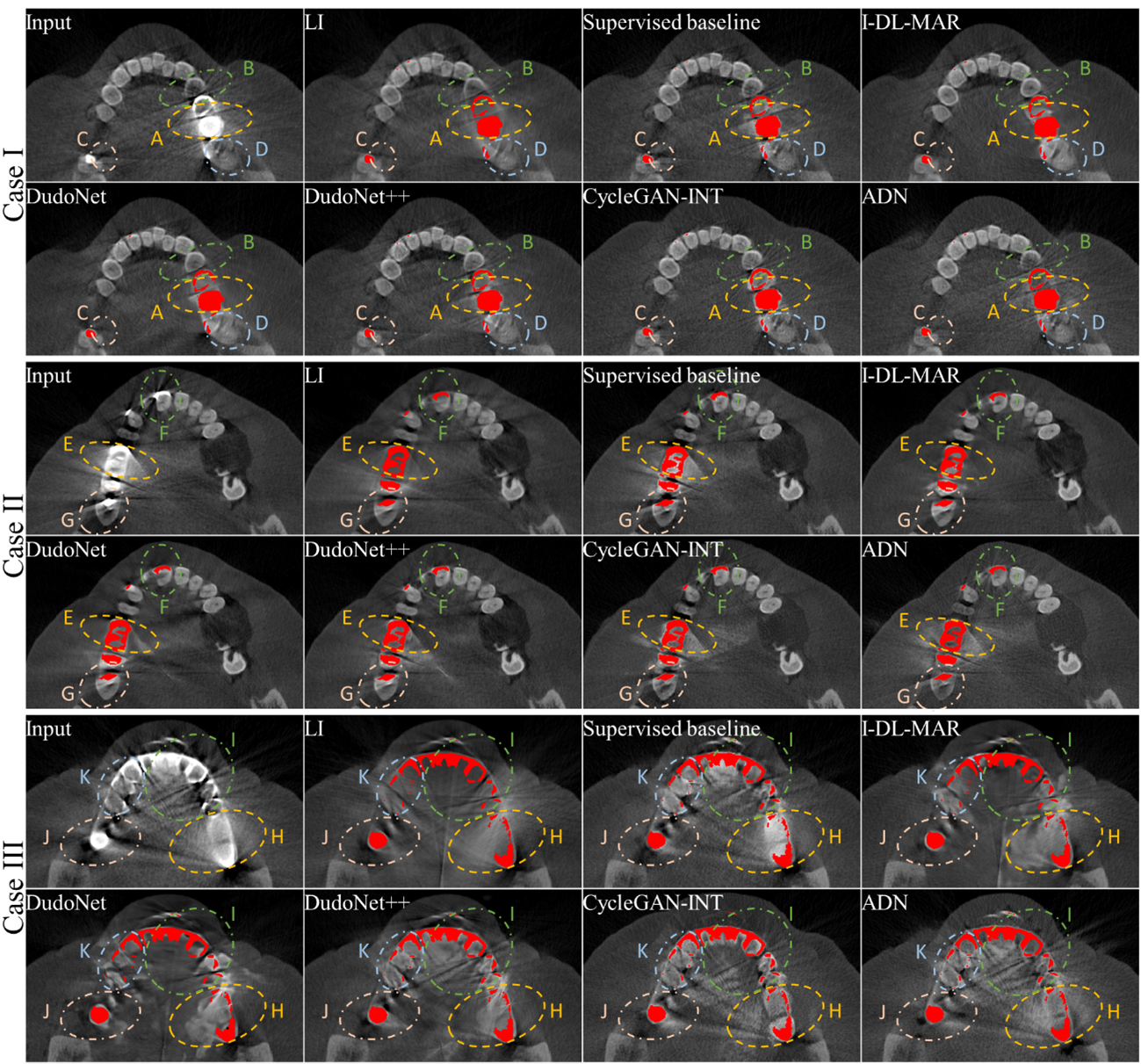}
    \caption{Visual results of three example clinical cases in dental dataset. There are multiple regions labeled for focus.}
    \label{fig:target_dental}
\end{figure}

\begin{figure}
    \centering
    \includegraphics{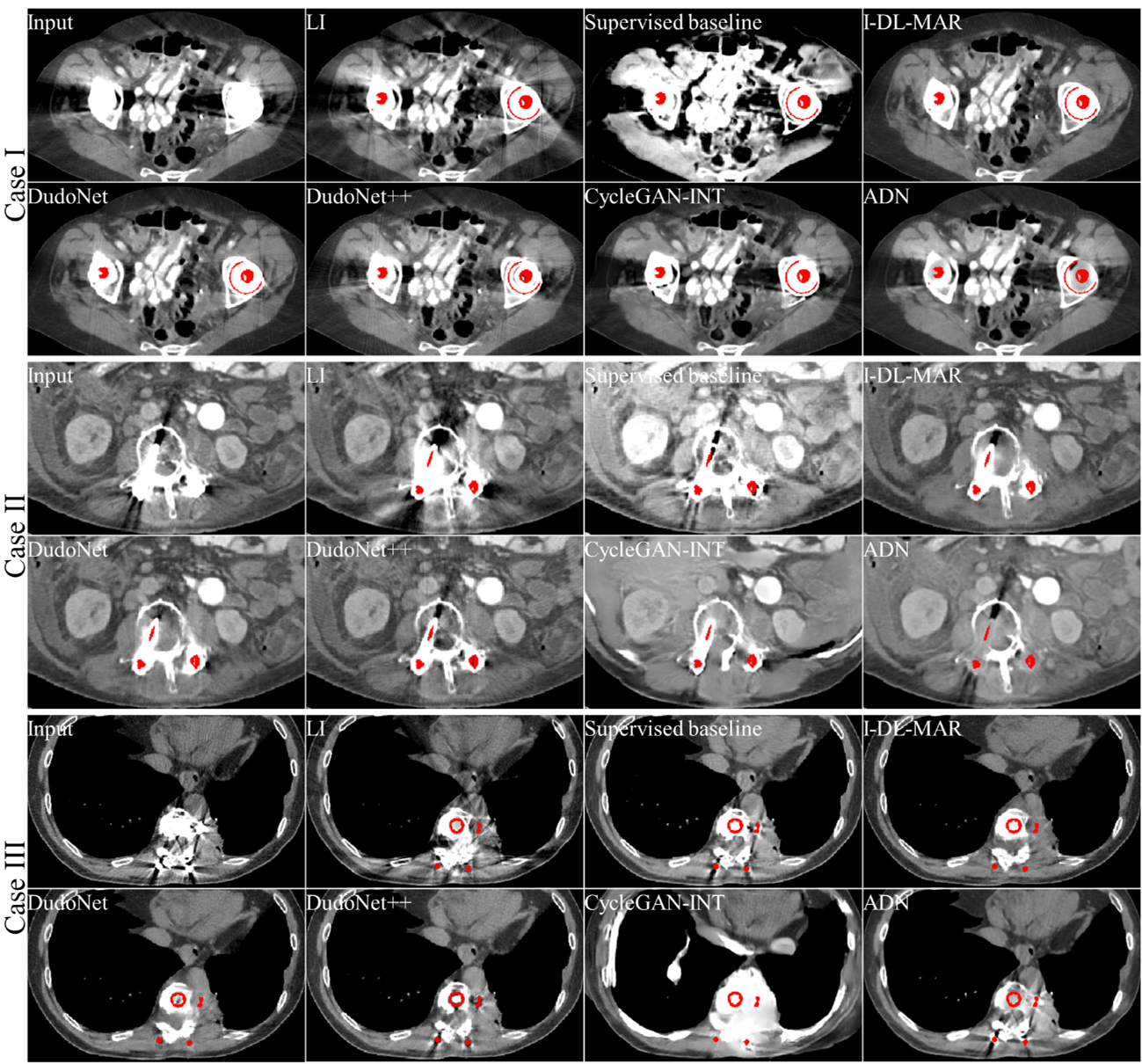}
    \caption{Visual results of three example clinical cases in torso dataset.}
    \label{fig:target_torso}
\end{figure}

\subsubsection{Dental dataset}
Figure \ref{fig:target_dental} shows three clinical cases in dental dataset. To facilitate the description of the results, several regions in each case are labeled. There are both solid metal implants, such as dental implants or the top of dental crowns, as shown in region C of Case I and region J of Case III, and hollow metal implants wrapped around the tooth, such as the middle of dental crowns, as shown in region A of Case I, region E of Case II and region I, K, H of Case III. From Case I to Case III, the number of metal implants increases, and the shape of the segmented metal shape becomes more complicated, leading to more and more severe metal artefacts. Among the labeled regions, region D and K mainly focuses on the ability to retain tooth or bone structure, while other regions mainly focus on the removal of artefacts. 

From Figure \ref{fig:target_dental}, we can see that, simple conventional method LI is partially effective to reduce artefacts in a few regions including B, C and F, but fails in other regions. Discarding information during LI also leads to new artefacts, as shown in regions H. Besides, structural loss happens in regions B, D, F and K.

Supervised baseline fails to reduce artefacts in almost all regions, indicating the existence of domain gap. However, it maintains the diagnostic structures nicely, without introducing significant blurring or distortion.

I-DL-MAR further enhances the ability of LI to remove artefacts from solid metal implants. It achieves best performance among all methods in Case I and Case II by reducing artefacts in most regions. However, it may be interfered by the new artefacts introduced in LI, as shown in region E of Case II and region K, H of Case III. Like LI, I-DL-MAR also leads to blurring or distorting diagnostic structures, as shown in region D of Case I and region K of Case III, although the problem has already been suppressed compared to LI.

For dual domain methods, on one hand, new artefacts are introduced in region E of Case II and region H of Case III, similar to situation in LI and I-DL-MAR. On the other hand, in Case I and certain regions in Case II and Case III, DudoNet and I-DL-MAR have similar behavior, while DudoNet++ and the supervised baseline behaves closely. Overall, the unsatisfying performance indicates domain gap also happens to dual-domain methods. The new artefacts happened in LI, I-DL-MAR and dual-domain methods may be related to their common feature of utilizing projection completion (by interpolation or CNN). The similar behavior of reducing artefacts between DudoNet and I-DL-MAR, and between DudoNet++ and the supervised baseline may be related to the kind of input they use (LI data or the original data).

The unsupervised methods ADN and CycleGAN-INT is comparable with the supervised baseline in reducing artefacts, even if the supervised baseline never sees target domain data during training. They are only able to reduce weak artefacts. Moreover, CycleGAN-INT faces another problem of changing diagnostic structure, as in Case I where many teeth are modified, or in region A of Case III, where areas destroyed by artefacts are replaced with implausible teeth. The results suggest unsupervised methods are not a satisfying solution to domain gap problem at least in our dental dataset.

\subsubsection{Torso dataset}
Figure \ref{fig:target_torso} shows three clinical cases in torso dataset. LI can reduce artefacts to some extent (e.g., Case III), but it can produce more severe new artefacts, which is similar to situation on source domain. The best methods are I-DL-MAR and DudoNet which further improve the results of LI by removing almost all the interpolation-based artefacts. The supervised baseline is only able to slightly reduce artefacts. However, the structures on various locations of the results are over-brightening or over-darkening. Such  anomalous behavior implies severe domain gap. DudoNet++ does not have the severe error problem as in the supervised baseline, and it is able to reduce more artefacts. However, it is less effective than I-DL-MAR and DudoNet. 

ADN is comparable to the supervised baseline. It can only remove thin streaking artefacts in torso dataset (see Case I and Case III) and slightly reduce dark banding artefacts. CycleGAN-INT can reduce more artefacts than ADN, however, the diagnostic structures change a lot and are untrustworthy.

Overall, for dental dataset, domain gap exists in all supervised methods. For torso dataset, domain gap exists mainly in methods taking the original data as input, including the supervised baseline and DudoNet++. By taking LI as input and learning to reduce interpolation-based artefacts, I-DL-MAR and DudoNet has the best visual performance in reducing artefacts, but they have problem of structural loss. Image-domain unsupervised methods trained on target domain is not able to reduce practical metal artefacts effectively  in neither dental dataset nor torso dataset, and therefore are not a satisfactory solution to the domain gap problem.

\subsection{Study of the impact of two input channels of IE-Net on the prediction of DudoNet and DudoNet++} \label{sec:IEExp}
In DudoNet and DudoNet++, domain gap problem is focused on the final prediction ${\hat\upmu}^\mathrm{fr,\ IE}$ of IE-Net, which has two input channels. One input channel is the reconstruction of SE-Net prediction ${\hat\upmu}^\mathrm{fr,\ SE}$, the other is the corrupted input image $\upmu$, which is LI image ${\upmu}^\mathrm{LI}$ in DudoNet or MA image ${\upmu}^\mathrm{ma}$ in DudoNet++. As IE-Net has two input channels, the relationship between the input and the domain gap of the final prediction is not as obvious as in the one-channel input image-domain MAR methods. Therefore, in order to investigate the cause of domain gap in dual-domain methods, which is discussed later in Section \ref{sec:domaingapDD}, it is important to understand the impact of two input channels of IE-Net on the final prediction ${\hat\upmu}^\mathrm{fr,\ IE}$, especially in terms of artefacts.

Generally, there might be artefacts in both ${\hat\upmu}^\mathrm{fr,\ SE}$ and $\upmu$, and each might influence artefacts in ${\hat\upmu}^\mathrm{fr,\ IE}$. To study the influence of channel A, the influence of the other channel B must be eliminated, which is equal to removing artefacts contained in channel B. In our experiment, it is achieved by replacing image in channel B with an image where metal artefacts have been reduced as much as possible. We call the new input image as pseudo-label, notated with $\upmu^\mathrm{p-fr}$. For source domain, we simply use the ground-truth artefact-free label as $\upmu^\mathrm{p-fr}$. For target domain where no paired labels exist, we use the image generated by the experimentally best MAR method as $\upmu^\mathrm{p-fr}$. In torso dataset, we choose I-DL-MAR to generate $\upmu^\mathrm{p-fr}$ according to the clinical study. While in dental dataset, as no method investigated in the clinical study has promising performance of MAR, we use a method that we are currently working on to generate $\upmu^\mathrm{p-fr}$. Although the chosen $\upmu^\mathrm{p-fr}$ in dental dataset, shown in Figure 5, still contains few artefacts, it is better than the results of all the investigated methods shown in shown in Figure \ref{fig:target_dental}.

\begin{figure}
    \centering
    \includegraphics{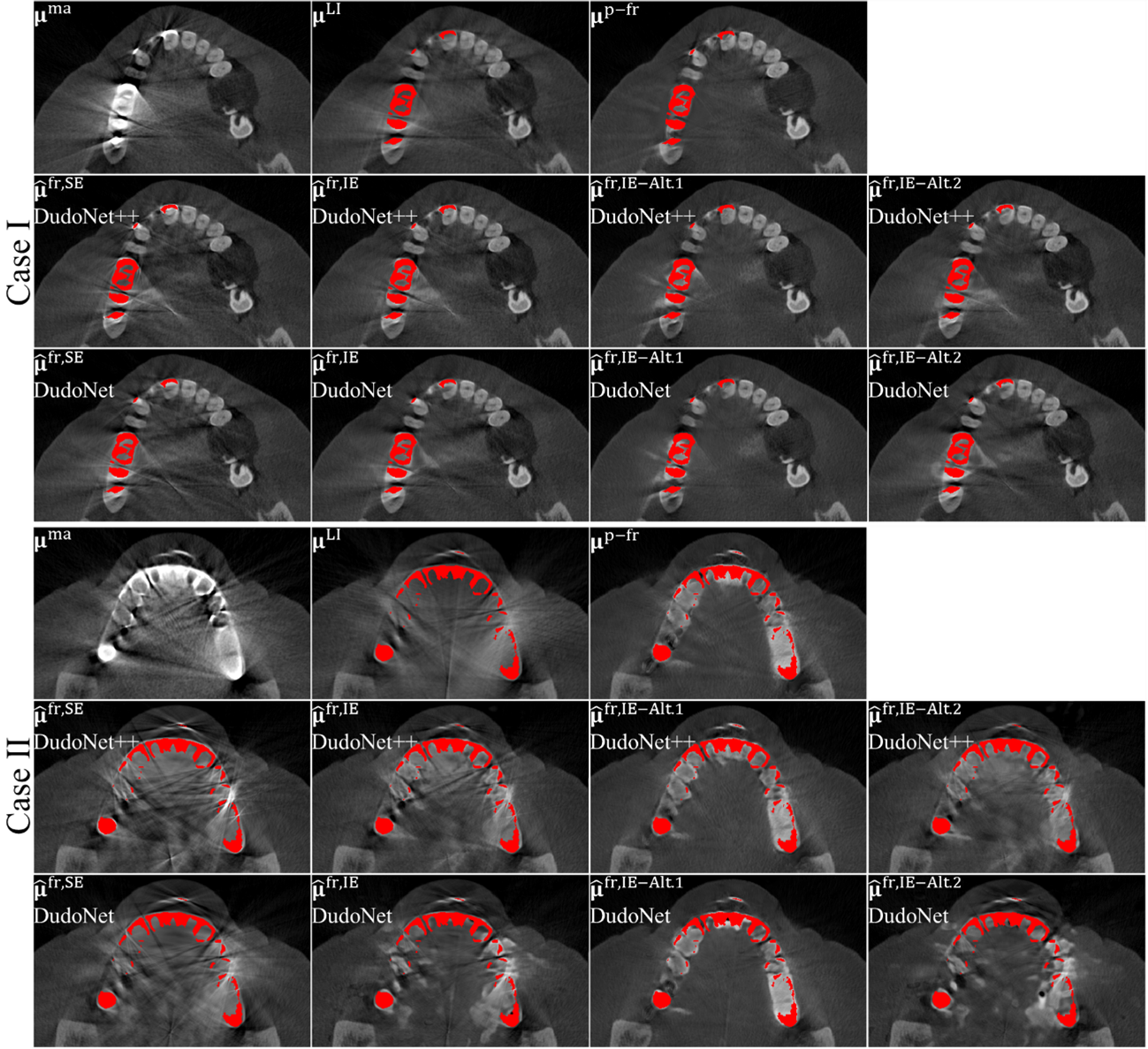}
    \caption{Visual results of two example practical cases in dental dataset, showing replacing one of the input channels of IE-Net with pseudo-label in dual-domain methods during testing. The first row for each case contains three types of input images including MA image, LI image and pseudo-label image. The second and third row contain the reconstruction of SE-Net prediction and three versions of IE-Net prediction, where the second row is for DudoNet++, and the third row is for DudoNet.}
    \label{fig:IEdental}
\end{figure}

\begin{figure}
    \centering
    \includegraphics{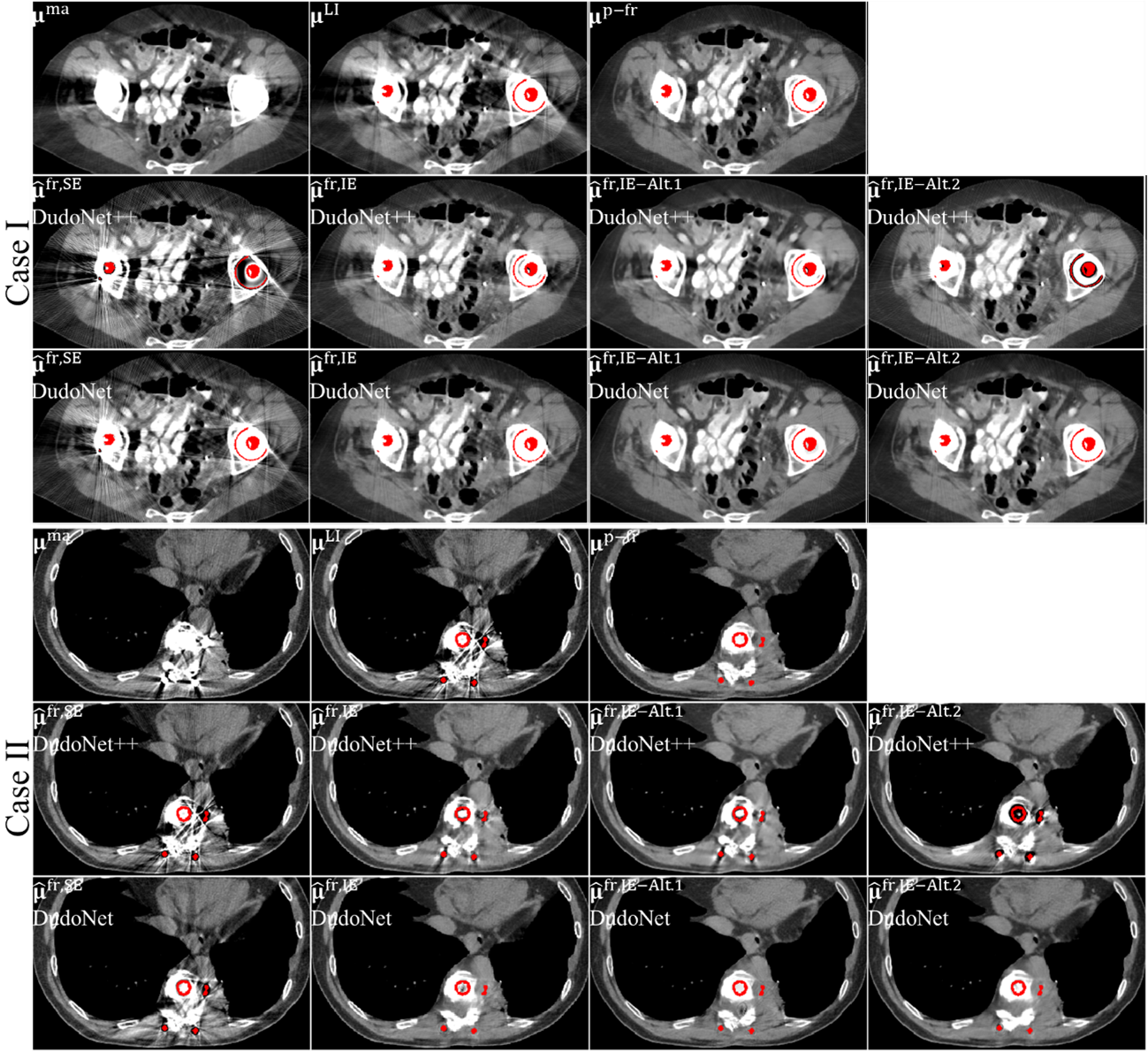}
    \caption{Visual results of two example practical cases in torso dataset, showing replacing one of the input channels of IE-Net with pseudo-label in dual-domain methods during testing. The first row for each case contains three types of input images including MA image, LI image and pseudo-label image. The second and third row contain the reconstruction of SE-Net prediction and three versions of IE-Net prediction, where the second row is for DudoNet++, and the third row is for DudoNet.}
    \label{fig:IEtorso}
\end{figure}

We notate the original IE-Net prediction and two alternative versions as follow:
\begin{itemize}
	\item The original IE-Net prediction whose input channels are ${\hat\upmu}^\mathrm{fr,\ SE}$ and $\upmu$ is notated as ${\hat\upmu}^\mathrm{fr,\ IE}$.
	\item The IE-Net prediction whose input channels are $\upmu^\mathrm{p-fr}$ and $\upmu$ is notated as ${\hat\upmu}^\mathrm{fr,\ IE-Alt1}$. 
	\item The IE-Net prediction whose input channels are ${\hat\upmu}^\mathrm{fr,\ SE}$ and $\upmu^\mathrm{p-fr}$ is notated as ${\hat\upmu}^\mathrm{fr,\ IE-Alt2}$.
\end{itemize}
The three versions are compared on both dental and torso datasets. The quantitative result on source domain is in Table \ref{tab:alternativeIENet} and the visual result on target domain of two datasets are presented in Figure \ref{fig:IEdental} and Figure \ref{fig:IEtorso}.

\begin{table}
	\caption{Quantitative evaluation results on source domain, with the input channel of IE-Net alternated in dual-domain methods for: (a) dental dataset and (b) Torso dataset. The best metric among the four results in each method is shown in bold.}
	\centering
	\resizebox{\linewidth}{!}{
	\renewcommand{\arraystretch}{1.3}
	\begin{tabular}{cc|cccc|cccc}
        \toprule
        \multicolumn{2}{c}{ \multirow{2}*{} } &\multicolumn{4}{c|}{DudoNet} &\multicolumn{4}{c}{DudoNet++}\\
        \cline{3-10}
        \multicolumn{2}{c}{} &${\hat\upmu}^\mathrm{fr,\ SE}$ & ${\hat\upmu}^\mathrm{fr,\ IE}$ & ${\hat\upmu}^\mathrm{fr,\ IE-Alt1}$ & ${\hat\upmu}^\mathrm{fr,\ IE-Alt2}$ &${\hat\upmu}^\mathrm{fr,\ SE}$ & ${\hat\upmu}^\mathrm{fr,\ IE}$ & ${\hat\upmu}^\mathrm{fr,\ IE-Alt1}$ & ${\hat\upmu}^\mathrm{fr,\ IE-Alt2}$\\
        \hline
        \multirow{3}*{\rotatebox{90}{(a) dental}}
        &RRMSE & 0.0643 & 0.0549 & \textbf{0.0223} & 0.0557 & 0.0565 & 0.0422 & \textbf{0.0226} & 0.0435\\
        &SSIM & 0.9638 & 0.9685 & \textbf{0.9971} & 0.9721 & 0.9692 & 0.9743 & \textbf{0.9968} & 0.9777\\
        &PSNR & 38.5199 & 39.4394 & \textbf{47.3192} & 39.3659 & 39.6503 & 41.6522 & \textbf{47.0976} & 41.3417\\
        \hline
        \multirow{3}*{\rotatebox{90}{(b) torso}}
        &RRMSE & 0.0595 & 0.0269 & 0.0307 & \textbf{0.0256} & 0.0724 & \textbf{0.0218} & \textbf{0.0218} & 0.1398\\
        &SSIM & 0.9635 & 0.9925 & 0.9926 & \textbf{0.9937} & 0.9519 & 0.9932 & \textbf{0.9943} & 0.8619\\
        &PSNR & 36.9789 & 44.7335 & 42.6043 & \textbf{45.0646} & 35.5252 & 46.0184 & \textbf{46.0499} & 30.0397\\
        \bottomrule
    \end{tabular}
    }
	\label{tab:alternativeIENet}
\end{table}

\subsubsection{Comparison of three versions of IE-Net in dental dataset}
On source domain, it can be seen from Table \ref{tab:alternativeIENet} (a) that for both DudoNet and DudoNet++, ${\hat\upmu}^\mathrm{fr,\ IE-Alt1}$ achieves significantly lower error than ${\hat\upmu}^\mathrm{fr,\ IE}$, while ${\hat\upmu}^\mathrm{fr,\ IE-Alt2}$ is close to ${\hat\upmu}^\mathrm{fr,\ IE}$. In other words, removing artefacts in ${\hat\upmu}^\mathrm{fr,\ SE}$ significantly improves the performance, while removing artefacts in $\upmu$ has little impact. This indicates that on source domain, the results of both DudoNet and DudoNet++ depend mainly on the reconstructed prediction of SE-Net ${\hat\upmu}^\mathrm{fr,\ SE}$, and are less influenced by the corrupted Input image $\upmu$.

On target domain, from Figure \ref{fig:IEdental}, we can firstly confirm that $\upmu^\mathrm{p-fr}$ contains far less artefacts than ${\upmu}^\mathrm{ma}$ or ${\upmu}^\mathrm{LI}$. For ${\hat\upmu}^\mathrm{fr,\ IE-Alt1}$ which replaces ${\hat\upmu}^\mathrm{fr,\ SE}$ with $\upmu^\mathrm{p-fr}$, the result is much better than ${\hat\upmu}^\mathrm{fr,\ IE}$, and is close to $\upmu^\mathrm{p-fr}$. Especially, in Case II, ${\hat\upmu}^\mathrm{fr,\ IE-Alt1}$ is even able to remove some artefacts remained in $\upmu^\mathrm{p-fr}$. On the other hand, for ${\hat\upmu}^\mathrm{fr,\ IE-Alt2}$ which replaces $\upmu$ with $\upmu^\mathrm{p-fr}$, the result is similar to ${\hat\upmu}^\mathrm{fr,\ IE}$. Both ${\hat\upmu}^\mathrm{fr,\ IE}$ and ${\hat\upmu}^\mathrm{fr,\ IE-Alt2}$ contain artefacts only found in ${\hat\upmu}^\mathrm{fr,\ SE}$ rather than $\upmu^\mathrm{p-fr}$. Above observation indicates on target domain, the results ${\hat\upmu}^\mathrm{fr,\ IE}$ of both DudoNet and DudoNet++ again mainly rely on ${\hat\upmu}^\mathrm{fr,\ SE}$ and are affected very little by $\upmu$.

Combining analysis on both source and target domain of the dental dataset, it can be concluded that for IE-Net in both DudoNet and DudoNet++, the input channel ${\hat\upmu}^\mathrm{fr,\ SE}$ dominates the final prediction, while the other input channel $\upmu$ has a limited affect.

\subsubsection{Comparison of three versions of IE-Net in torso dataset}
On source domain, we can see from Table \ref{tab:alternativeIENet} (b) that, for DudoNet, ${\hat\upmu}^\mathrm{fr,\ IE-Alt1}$ and ${\hat\upmu}^\mathrm{fr,\ IE}$ perform similarly, while ${\hat\upmu}^\mathrm{fr,\ IE-Alt2}$ outperforms the others by a small margin.

For DudoNet++, ${\hat\upmu}^\mathrm{fr,\ IE-Alt1}$ is slightly better than ${\hat\upmu}^\mathrm{fr,\ IE}$, while ${\hat\upmu}^\mathrm{fr,\ IE-Alt2}$ has much higher error than the others. Specially, we found that SE-Net in DudoNet++ performs poorly in reducing metal artefacts, and there are overlapping regions of artefacts in ${\hat\upmu}^\mathrm{fr,\ SE}$ and $\upmu$, as shown in Figure 6. When the original IE-Net was trained to remove the artefacts in the combination of ${\hat\upmu}^\mathrm{fr,\ SE}$ and $\upmu$, it might learn an empirical and structure-related rule for modification, such as recognizing and lightening certain regions containing dark banding or shading artefacts. By replacing ${\hat\upmu}^\mathrm{fr,\ SE}$ with $\upmu^\mathrm{p-fr}$, IE-Net might still modify certain regions with a similar degree, resulting in over-brightness. The problem does not occur in DudoNet, probably because SE-Net in DudoNet reduces dark banding and shading artefacts better than SE-Net in DudoNet++, thus IE-Net learns different rules that do not lead to such problem.

In above results, replacing neither channel with $\upmu^\mathrm{p-fr}$ would result in significantly lower error, thus suggesting that IE-Net in DudoNet and DudoNet++ is not simply dominated by any single input channel. ${\hat\upmu}^\mathrm{fr,\ SE}$ and $\upmu$ might jointly influence the final output, although the mechanisms behind it might be different in DudoNet and DudoNet++.

On target domain, former clinical study has shown that in torso dataset, DudoNet is affected less by domain gap, while DudoNet++ is affected more easily. In this experiment, we found from Figure 6 that:

For DudoNet, ${\hat\upmu}^\mathrm{fr,\ SE}$ and ${\upmu}^\mathrm{LI}$ both contains certain degree of artefacts, but ${\hat\upmu}^\mathrm{fr,\ IE}$ can reduce the majority of artefacts, with a few streaking artefacts remained. By reducing artefacts in one input channel, ${\hat\upmu}^\mathrm{fr,\ IE-Alt1}$ and ${\hat\upmu}^\mathrm{fr,\ IE-Alt2}$ both reduce artefacts further and are closer to $\upmu^\mathrm{p-fr}$. The results suggest that both input channels to IE-Net in DudoNet affect the final prediction.

For DudoNet++, it is interesting to observe that the remained artefacts in the three versions of IE-Net prediction are a fused preservation of artefacts in corresponding two input channels. For example, in Case I, among candidates for input channels to IE-Net, ${\upmu}^\mathrm{ma}$ contains relatively more dark banding and shading artefacts, ${\hat\upmu}^\mathrm{fr,\ SE}$ contains relatively more streaking artefacts, while $\upmu^\mathrm{p-fr}$ contains few artefacts. Among the predictions, ${\hat\upmu}^\mathrm{fr,\ IE}$ contains both dark banding and shading artefacts from ${\upmu}^\mathrm{ma}$ and streaking artefacts from ${\hat\upmu}^\mathrm{fr,\ SE}$, ${\hat\upmu}^\mathrm{fr,\ IE-Alt1}$ mainly contains dark banding and shading artefacts from ${\upmu}^\mathrm{ma}$, while ${\hat\upmu}^\mathrm{fr,\ IE-Alt2}$ only contains streaking artefacts from ${\hat\upmu}^\mathrm{fr,\ SE}$,  even if there are also dark banding artefacts in ${\hat\upmu}^\mathrm{fr,\ SE}$. Similar phenomenon also occurs in Case II. The results suggest that the artefacts in the prediction of DudoNet++ comes from both input channels to IE-Net. Specially, the remaining artefacts are not a simple combination of all artefacts in two input channels, but a more complex process that can be observed case-by-case but hard to be accurately modeled.

Combining above analysis on source and target domain, it can be concluded that for DudoNet and DudoNet++, ${\hat\upmu}^\mathrm{fr,\ IE}$ and $\upmu$ jointly determine the final prediction. Neither input channel is dominant, and the process to fuse and reduce artefacts from two channels is complex. 

\section{Discussion}
\subsection{Probable causes of domain gap problem in supervised MAR methods}
In above section, we have conducted simulation and clinical studies in dental dataset and torso dataset. By comparing results in two studies, we observe that all supervised MAR methods investigated here are troubled with domain gap problem to some extent. Specifically, in dental dataset, all the supervised methods perform well on source domain, but are unsatisfying on target domain. While in torso dataset, I-DL-MAR and DudoNet are less prone to domain gap problem, the supervised baseline meets severe problem, DudoNet++ meets mild problem. It is worth discussing where domain gap problem comes from and why domain gap problem varies for different combinations of methods and datasets.
In this section, we try to answer above questions. Specifically, we summarize the investigated methods to image-domain method and dual-domain method and discuss them separately. The dual-domain method contains a sinogram-domain SE-Net and an image-domain IE-Net. IE-Net here is different from common image-domain supervised methods like the supervised baseline and I-DL-MAR, as IE-Net is two-channel-input while the common ones are one-channel-input. Thanks to the experimental study and analysis in Section \ref{sec:IEExp}, we have shown that how the prediction of IE-Net is related to the two input channels. Therefore, we can safely discuss only the sinogram-domain method and one-channel-input image-domain method. The cause of domain gap in dual-domain methods can be completed by combining above conclusion.

\subsubsection{Sinogram-domain method: SE-Net of DudoNet and DudoNet++} \label{sec:domaingapSE}

\begin{figure}
    \centering
    \includegraphics{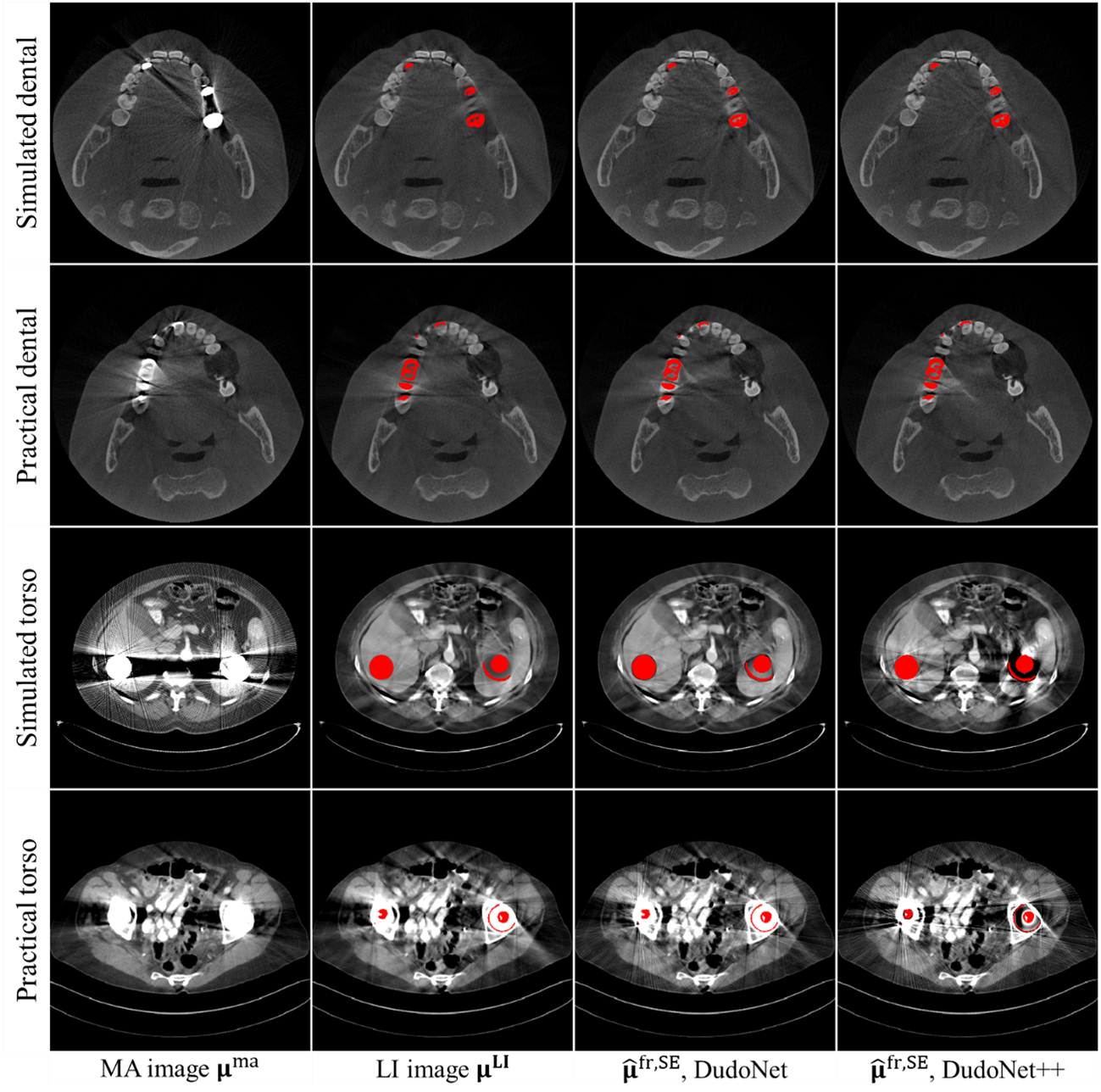}
    \caption{Reconstruction of the SE-Net prediction ${\hat\upmu}^\mathrm{fr,\ SE}$ of DudoNet and DudoNet++ on four types of data, shown together with ${\upmu}^\mathrm{ma}$ and ${\upmu}^\mathrm{LI}$.}
    \label{fig:SE}
\end{figure}

For sinogram-domain methods, we take SE-Net of DudoNet and DudoNet++ as examples. Other sinogram-domain methods can be analyzed in a similar approach.
\paragraph{Existence of domain gap}
Figure \ref{fig:SE} shows the example results of the reconstructed prediction of SE-Net ${\hat\upmu}^\mathrm{fr,\ SE}$ in DudoNet and DudoNet++ on both source domain and target domain of the two datasets. The cases here all appear in previous simulation and clinical studies reconstruction, thus readers can refer to Figure 1~4 to check the difference.

In dental dataset, SE-Net in both methods perform relatively well on source domain but new or remained artefacts exist on target domain, indicating that SE-Net has domain gap problem. While in torso dataset, similar domain gap happens for SE-Net of both methods. It is noticeable that in the exemplar cases, not only MA image appears differently on source and target domain, but LI image also appears differently. Moreover, the metal mask on target domain covers a smaller percentage of suspicious metal area than on source domain.
\paragraph{Cause of domain gap}

SE-Net is input with two elements: the corrupted sinogram $\mathbf{p}$, which are LI projection $\mathbf{p}^\mathrm{LI}$ for DudoNet or the original artefact-affected projection $\mathbf{p}^\mathrm{ma}$ for DudoNet++, and the metal trace binary mask $\mathbf{M}_t\left(\mathbf{p}^{\mathrm{ma}}\right)$. SE-Net in both methods only modifies $\mathbf{p}$ merely inside the metal trace. Therefore, for the output of SE-Net, domain gap problem may come from both the sinogram and the metal trace.

\begin{itemize}
	\item \textbf{Sinogram}
	
The mismatch between acquisition of artefact-affected CT projections in simulation and practice include many aspects, such as detector response, forward model, source spectrum and noise. Thus, there is unavoidable mismatching between projections in simulation and practice, and domain difference exists in the original metal-artefact-affected projections  $\mathbf{p}^\mathrm{ma}$. While for LI projections $\mathbf{p}^\mathrm{LI}$, signals inside the metal trace are replaced with the linear interpolation of the signals at the outer contour of the metal trace. Only if the metal trace covers the regions affected by metals correctly, linear interpolated signals can be theoretically metal-free and thus domain-invariant. As a result, for SE-Net in DudoNet++, using  $\mathbf{p}^\mathrm{ma}$ brings in domain gap, while for SE-Net in DudoNet, using $\mathbf{p}^\mathrm{LI}$ generated from correct metal trace will not lead to domain gap.

However, in practice, it might be hard to acquire correct metal trace, as will be discussed later. Using metal trace that does not cover the entire corrupted region (often seen in under-segmentation situation) will result in metal-affected signals leaked into LI signals, which brings in domain-variant information. Using metal trace that covers regions including too much normal signals (often seen in over-segmentation situation) will result in information loss in projection and structural loss in reconstructed image. If $\mathbf{p}^\mathrm{LI}$ on target domain meets at least one of the above situations, then $\mathbf{p}^\mathrm{LI}$ on target domain is different from $\mathbf{p}^\mathrm{LI}$ on source domain (generally generated with correct metal trace), thus domain gap is introduced.
	\item \textbf{Metal trace}
	
We have mentioned that metal trace is an input element to SE-Net and plays a key role in generating $\mathbf{p}^\mathrm{LI}$. Thus, if domain gap in metal trace exists, the ability of SE-Net to generalize on target domain may degrade.

The inaccurate segmentation of metal trace has been discussed earlier, usually appearing as at least one of under-segmentation and over-segmentation. Using inaccurate metal trace on target domain will result in domain gap. Besides, as SE-Net only modifies projection inside the metal trace, using inaccurate metal trace on target domain means unnecessary or missing modification and increasing error after reconstruction.

Another cause of domain gap is the unmatched distribution of shape and complexity of metals between source and target domain. As metal trace is input to SE-Net, the unmatched distribution of metals will result in unmatched distribution of metal traces and domain gap.

In experiments, domain gap problem of metal trace depends on datasets.

In dental dataset, both the inaccurate metal trace segmentation and mismatched distribution of metal traces are prominent. In terms of the accuracy of segmentation, on source domain, it is easier to segment metals. On target domain, since metals and teeth all have high attenuation, the strong artefacts may interfere with the boundary between metals and teeth in the reconstruction image. Thus, segmentation of metals becomes more difficult and less accurate. In terms of the distribution of metal traces, it is related to the distribution of inserted metals. In practice, metals in the dental CT images are of various shapes like the middle intersection of dental crowns, dental braces, and dental implants. While in the simulation, the inserted metals are more like simpler dental implants or the top intersection of the dental crowns, which are less complex than some practical cases, making the distribution of metals mismatched. Thus, domain gap is prominent in dental dataset.

In torso dataset, due to the lower attenuation of materials in the human body compared to dental cases, the artefacts are less severe, so metals are easier to segment. Besides, as the accuracy of segmenting metal trace is mathced in source and target domain, and the distribution of the shape of metal trace in source domain is closer to the distribution in target domain, domain difference in segmenting metal traces is less pronounced.

\end{itemize}

\subsubsection{One-channel-input image-domain method: the supervised baseline and I-DL-MAR}
For one-channel input image-domain networks, we take the supervised baseline and I-DL-MAR as examples.

\paragraph{Existence of domain gap}
Comparing results in simulation study and clinical study, it is clear to see that domain gap in the supervised baseline exists in both dental and torso dataset, while domain gap in I-DL-MAR mainly exists in dental dataset and is hardly seen in torso dataset.
\paragraph{Causes of domain gap}
For one-channel input image-domain networks, domain gap mainly comes from the domain-variant input image. 
For the supervised baseline, the input is the original artefact-affected image ${\upmu}^\mathrm{ma}$, whose metal artefacts are theoretically different between simulation and practice in any dataset. The domain difference of ${\upmu}^\mathrm{ma}$ mainly comes from the domain difference in $\mathbf{p}^\mathrm{ma}$. In our experiments, it is observed in both dental and torso datasets, explaining domain gap in the supervised baseline.
For I-DL-MAR, the input is LI image ${\upmu}^\mathrm{LI}$, which is reconstructed from LI projection $\mathbf{p}^\mathrm{LI}$. The existence and causes of domain difference of $\mathbf{p}^\mathrm{LI}$ have been discussed before. Similarly, domain difference of ${\upmu}^\mathrm{LI}$ mainly exists in dental dataset, explaining domain gap in I-DL-MAR.

\subsubsection{Dual-domain method: DudoNet and DudoNet++}
\label{sec:domaingapDD}
\paragraph{Existence of domain gap}
From experimental results, we can observe that domain gap  in DudoNet mainly exists in dental dataset, while domain gap in DudoNet++ exists in both dental and torso dataset.
\paragraph{Causes of domain gap}
The causes of domain gap in dual-domain methods can be deduced from the analysis of two-channel-input IE-Net in Section \ref{sec:IEExp} and the probable causes of domain gap in SE-Net in Section \ref{sec:domaingapSE}.

For DudoNet, in dental dataset, as the result of IE-Net is mainly determined by the input channel ${\hat\upmu}^\mathrm{fr,\ SE}$ but not $\upmu$, the cause of domain gap is the same with SE-Net, which is the metal trace $\mathrm{M}_t(\mathbf{p}^{\mathrm{ma}})$ and the LI sinogram $\mathbf{p}^\mathrm{LI}$. While in torso dataset, DudoNet meets nearly no domain gap problem.
For DudoNet++, in dental dataset, similar to DudoNet, domain gap comes from the prediction of SE-Net, which is related to the metal trace $\mathrm{M}_t(\mathbf{p}^{\mathrm{ma}})$ and the original sinogram $\mathbf{p}^\mathrm{ma}$. While in torso dataset, the result of IE-Net is affected by both two input channels, the domain gap comes from both $\upmu^{\mathrm{ma}}$ and ${\hat\upmu}^\mathrm{fr,\ SE}$ which is in turn related to $\mathbf{p}^\mathrm{ma}$. The metal trace is nearly domain-invariant in torso dataset, thus is ignored to be the cause.

In conclusion, the causes of domain gap in methods investigated in two datasets are summarized in Table \ref{tab:domaingapcause}. It only considers the factor that is directly input to the method and contains domain difference. For example, for I-DL-MAR, we only show the cause of domain gap in dental dataset because domain gap in torso dataset is not obvious. In addition, we only consider the influence of ${\upmu}^\mathrm{LI}$ instead of $\mathbf{p}^\mathrm{LI}$ and $\mathbf{M}_t\left(\mathbf{p}^{\mathrm{ma}}\right)$ because the latter two are not directly input to I-DL-MAR.

\begin{table}[]
    \caption{The causes of domain gap in supervised methods in: (a) dental dataset and (b) Torso dataset. ID means image-domain, DD means dual-domain.}
    \centering
    \begin{tabular}{cc|ccccc|ccccc}
        \toprule
        \multirow{2}*{} & {}
        &\multicolumn{5}{c|}{(a) dental} &\multicolumn{5}{c}{(b) torso}  \\
        \cline{3-12}
        &{\diagbox{Methods}{Causes}} &{$\mathbf{p}^{\mathrm{ma}}$} & {$\mathbf{p}^{\mathrm{LI}}$} & {$\mathrm{M}_t(\mathbf{p}^{\mathrm{ma}})$} & {$\upmu^{\mathrm{ma}}$}
        & {$\upmu^{\mathrm{LI}}$} &{$\mathbf{p}^{\mathrm{ma}}$} & {$\mathbf{p}^{\mathrm{LI}}$} & {$\mathrm{M}_t(\mathbf{p}^{\mathrm{ma}})$} & {$\upmu^{\mathrm{ma}}$}
        & {$\upmu^{\mathrm{LI}}$}\\
        \hline
        \multirow{2}*{ID}
         &  {Supervised baseline} &{} &{} &{} &{\checkmark} &{} &{} &{} &{} &{\checkmark} &{}\\
         &  {I-DL-MAR} &{} &{} &{} &{} &{\checkmark} &{} &{} &{} &{} &{} \\
        \hline
         \multirow{2}*{DD}
         &  {DudoNet} &{} &{\checkmark} &{\checkmark} &{} &{} &{} &{} &{} &{}\\
         &  {DudoNet++} &{\checkmark} &{} &{\checkmark} &{} &{} &{\checkmark} &{} &{} &{\checkmark} &{}\\
         \bottomrule
    \end{tabular}
    
    \label{tab:domaingapcause}
\end{table}

\subsection{Characteristics of investigated methods}

\subsubsection{The supervised baseline}
The supervised baseline is one of the simplest methods investigated in this work. It performs well on source domain but meets severe domain gap problem on target domain in both dental and torso dataset, thus is not suitable for practice.

\subsubsection{I-DL-MAR}
I-DL-MAR is as simple as the supervised baseline but shows much better performance on target domain. Domain gap problem in I-DL-MAR is closely related to the domain difference in LI image. I-DL-MAR can overcome domain gap problem only if the metal trace used in LI is correctly segmented and its distribution is domain-invariant. Such requirements are met in torso dataset, but not perfectly met in dental dataset, thus I-DL-MAR performs unsatisfying on target domain in dental dataset. Besides, another problem of I-DL-MAR is the structural loss near metal implants, which is caused by the information loss in LI. Although it manages to recover some of the distorted structures in LI image through data-driven learning, the recovery might be incomplete and sometimes unreliable.

In practice, to overcome domain gap problem of I-DL-MAR in clinical dental CT, two approaches should be taken:

1. During simulation, using metals with distribution as close to the clinical situation as possible, like increasing the number of types and complexity of metals.

2. During inference on target domain, improving the accuracy of metal segmentation, which is difficult sometimes and may require manual labeling.

\subsubsection{DudoNet}
As a dual-domain method that takes LI data as input, DudoNet is more complex compared to I-DL-MAR. It achieves better performance quantitatively on source domain, although the gain over I-DL-MAR is not obvious in visual comparison.
On target domain, DudoNet behaves similarly to I-DL-MAR. It meets the same domain gap problem in dental dataset, while performs well in torso dataset. The domain gap in dental dataset comes from domain difference in LI projection and metal trace, which is basically the same as I-DL-MAR. Therefore, in order to apply DudoNet in clinical dental CT, same approaches as for I-DL-MAR should be taken.
In addition, like I-DL-MAR, structural loss happens in the results of DudoNet, due to the use of LI data.

\subsubsection{DudoNet++}
DudoNet++ makes several modifications to DudoNet, among which the most important one is using the original artefact-affected data as input. From simulation study we can see that the modifications are worthy, as DudoNet++ achieves best performance on source domain both quantitatively and visually. Most importantly, it solves the structural loss happening in DudoNet.

However, on target domain, using the original data as input brings severe domain gap problem. In dental dataset, domain gap mainly comes from the original artefact-affected projection and the metal trace. In torso dataset, domain gap mainly comes from the original artefact-affected projection and its reconstruction image. Thus, even taking approaches to make the metal trace domain-invariant, DudoNet++ still cannot solve domain gap problem caused by the input data.

\subsubsection{CycleGAN-INT}
The advantage of unsupervised MAR methods is that they can be trained and tested on same data domain. Thus, ideally unsupervised MAR methods can avoid domain gap problem. However, unsupervised methods often have more complex designs than supervised methods, and the training might be difficult or unstable.

In our experiments, we find it hard to train CycleGAN-INT to perform well on either source domain or target domain. We have tried a large combinations of training hyperparameters and find that there is always a trade-off between the ability of MAR and the ability to keep diagnostic structures unchanged. Adding the intensity loss \citep{nakao_3dcyclegan_2020} only partially alleviates the problem. This might be due to the characteristic of CycleGAN which relies on GANs to learn the features of artefact-free images and artefact-affected images. Currently, for two datasets used in this work, CycleGAN-INT cannot give satisfying prediction.

\subsubsection{ADN}
Compared to CycleGAN based MAR, ADN designs a more complex mechanism to encourage the disentanglement of metal artefacts and artefact-free images. In our experiments, we find that for most cases ADN does not have the problem of changing structures as seen in CycleGAN-INT. However, its ability to reduce metal artefacts is close to CycleGAN-INT. On source domain, it is much worse than the supervised methods. On target domain, it can only reduce certain types of artefacts to some extent. Currently, for two datasets used in this work, ADN is not a satisfying MAR method to solve domain gap problem.

\section{Conclusion}
\label{sec:conclusion}
Recent years, there are more and more deep-learning-based methods proposed for metal artefact reduction in CT images. Supervised MAR methods trained on simulated data may perform worse on practical data, due to the difference between simulated and practical artefacts. Such phenomenon is the domain gap problem in CT MAR task. Several unsupervised MAR methods are proposed to address domain gap problem by directly learning on practical data. Until now, there lacks a comparison of mainstream deep MAR methods on solving domain gap problem.

In this work, we experimentally investigate domain gap problem in six mainstream deep MAR methods. The comparison is conducted in a dental dataset and a torso dataset. The investigated methods include image-domain supervised MAR (an image-domain supervised baseline and I-DL-MAR), dual-domain supervised MAR (DudoNet and DudoNet++) and image-domain unsupervised MAR (CycleGAN-based MAR and ADN).

Several findings are made from the experiments. For the supervised MAR methods, in the dental dataset all the methods have domain gap problem, while in the torso dataset I-DL-MAR and DudoNet are basically not troubled with domain gap problem. Generally, using linear-interpolated data as input can address domain gap problem to some extent, but it may result in structural loss. While using the original artefact-affected data as input is in opposite troubled with severe domain gap problem but almost no structural loss. For the unsupervised MAR methods, CycleGAN-based MAR and ADN cannot perform well on neither simulated artefacts nor practical artefacts in two datasets, thus are not satisfying solution to domain gap problem. New method to solve domain gap problem in CT MAR is needed, especially for the dental CT scenario.

We also analyze the probable causes of domain gap problem in the supervised methods, finding that actual causes vary with the method and the dataset. Generally, these causes can be summarized to the artefact-affected sinograms and images, as well as the metal trace in the sinograms. The analysis can be used to optimize existing methods or design new methods.

\bibliographystyle{unsrtnat}
\bibliography{domain-gap-mar}

\end{document}